\documentclass{bmvc2k}

\pdfoutput=1
\usepackage{hyperref}
\usepackage{cleveref}

\usepackage{amsmath,graphicx}
\usepackage{subfigure}
\usepackage{multirow}
\usepackage{amssymb}
\usepackage{enumitem}
\usepackage{dsfont}
\usepackage{amsfonts}
\newcommand{\eg}{e.g.}
\newcommand{\ie}{i.e.}
\newcommand{\cL}{\mathcal{L}}
\newcommand{\cR}{\mathcal{R}}
\newcommand{\cD}{\mathcal{D}}
\newcommand{\norm}[1]{{\| #1 \|}^2}

\def\eg{\emph{e.g}}

\title{Learning Generic Diffusion Processes for Image Restoration}

\addauthor{Peng Qiao}{pengqiao@nudt.edu.cn}{1}
\addauthor{Yong Dou}{yongdou@nudt.edu.cn}{1}
\addauthor{Yunjin Chen}{chenyunjin nudt@hotmail.com}{2}
\addauthor{Wensen Feng}{sanmumuren@126.com}{3}

\addinstitution{
 Science and Technology on Parallel and Distributed Laboratory\\
 National University of Defense Technology\\
 Changsha, China
}
\addinstitution{
 ULSee, Inc.\\
 Hangzhou, China
}
\addinstitution{
 College of Computer Science \& Software Engineering\\
 Shenzhen University\\
 Shenzhen, China
}

\runninghead{Q. Peng, Y. Dou, Y. Chen, W. Feng}{Learning Generic Diffusion Processes}

\def\eg{\emph{e.g}\bmvaOneDot}

\begin{document}

\maketitle

\begin{abstract}
Image restoration problems are typical ill-posed problems where the regularization term plays an important role.
The regularization term learned via generative approaches is easy to transfer to various image restoration, but offers inferior restoration quality compared with that learned via discriminative approaches.
On the contrary, the regularization term learned via discriminative approaches are usually trained for a specific image restoration problem, and fail in the problem for which it is not trained.
To address this issue, we propose a generic diffusion process (genericDP) to handle multiple Gaussian denoising problems based on the Trainable Non-linear Reaction Diffusion (TNRD) models.
Instead of one model, which consists of a diffusion and a reaction term, for one Gaussian denoising problem in TNRD, we enforce multiple TNRD models to share one diffusion term.
The trained genericDP model can provide both promising denoising performance and high training efficiency compared with the original TNRD models.
We also transfer the trained diffusion term to non-blind deconvolution which is unseen in the training phase.
Experiment results show that the trained diffusion term for multiple Gaussian denoising can be transferred to image non-blind deconvolution as an image prior and provide competitive performance.
\end{abstract}

\section{Introduction}
\label{sec:intro}
Image restoration problems, \eg, image denoising, deconvolution, super-resolution and et. al,
have been researched for decades, and are still active research areas.
In image restoration problems, we aim to recover the clean image $u$,
given its degraded counterpart $f$ generating by the following procedure,
\begin{equation}\label{eq:restorationProblem}
f = Au + v,
\end{equation}
where $v$ is the added noise, for Gaussian denoising, $v$ is assumed to be additive zero mean Gaussian noise.
$A$ is the degradation operator, \eg,
for image denoising, $A$ is identity matrix;
for image super-resolution, $A$ is decimating operator;
for image deconvolution, $A$ is blur operator.

\subsection{Related Works}
It is well known that image restoration problems are typical ill-posed problems.
Variational approaches are suitable to solve these problems with the proper regularization terms.
A typical variational model is given as
\begin{equation}\label{eq:regularizationModel}
E(u,f) = \mathcal{R}(u) + \mathcal{D}(u,f),
\end{equation}
where $\mathcal{R}(u)$ is the regularization term, and
$\mathcal{D}(u,f)$ is the data term.
Widely used image regularization term models include the most well-known
Total Variation (TV) functional \cite{rudin1992nonlinear},
Total Generalized Variation (TGV) \cite{bredies2010total},
Expected Patch Log Likelihood (EPLL) \cite{zoran2011epll} and
Fields of Experts (FoE) based analysis operator \cite{roth2005fields, chen2014insights}.

In recent years, machine learning based approaches have achieved better restoration performance compared with the hand-crafted regularization terms and widely used BM3D \cite{dabov2007image}.
The machine learning based approaches can be divided into two groups, generative approaches and discriminative approaches.
Generative approaches, \eg,
FoE \cite{roth2005fields},
K-SVD \cite{elad2006image} and
EPLL \cite{zoran2011epll},
aim to learn the probabilistic model of natural images,
which is used as the regularization term to recover various degraded images.
On the contrary, discriminative approaches aim to learn the inference procedure that minimizes the energy (\ref{eq:regularizationModel}) using pairs of degraded and clean images, \eg,
Cascade Shrinkage Fields (CSF, \cite{schmidt2014shrinkage}), and
Trainable Non-linear Reaction Diffusion (TNRD, \cite{chen2015learning}).

\subsection{Our Motivations and Contributions}
Taking TNRD as an example, while it offers both high computational efficiency and high restoration quality, it is highly specified for a specific restoration problem and fails in the problem for which it is not trained. If we have to handle Gaussian denoising problems of $\sigma$ ranging from 1 to 50, we need to train 50 models. In terms of training efficiency, the performance of discriminative approaches is not as good as that of generative approaches. To address this issue, we propose a generic diffusion process model to combine the advantages of both discriminative and generative approaches, \ie, training a model that provides high training efficiency and achieves competitive restoration quality. The contribution of this study is summarized as follows:

\textbf{(a)}
Unlike one model for one Gaussian denoising problem in TNRD, we enforce multiple TNRD models handling different noise level to share one diffusion term.

\textbf{(b)}
We give the derivations of the gradients for training the proposed model, which is important for training the proposed model.
We train the parameters in a supervised manner, and train the models in an end-to-end fashion.

\textbf{(c)}
We transfer the trained diffusion term to deal with non-blind image deconvolution which is unseen in the training phase.
The resulting optimization problem is optimized via the half quadratic splitting (HQS).

\textbf{(d)}
Experiment results show that the genericDP model can achieve almost the same performance compared with the TNRD model trained for a specific $\sigma$.
In non-blind image deconvolution problem, with this trained diffusion term, it provides competing results.

\section{Generic Diffusion Process}

In this section, we first introduce the proposed generic diffusion process (genericDP), and then we give the gradients of the loss function w.r.t. the parameters.

\subsection{Generic Diffusion Process}
\begin{figure}[t!]
\centering
\includegraphics[width=0.7\linewidth]{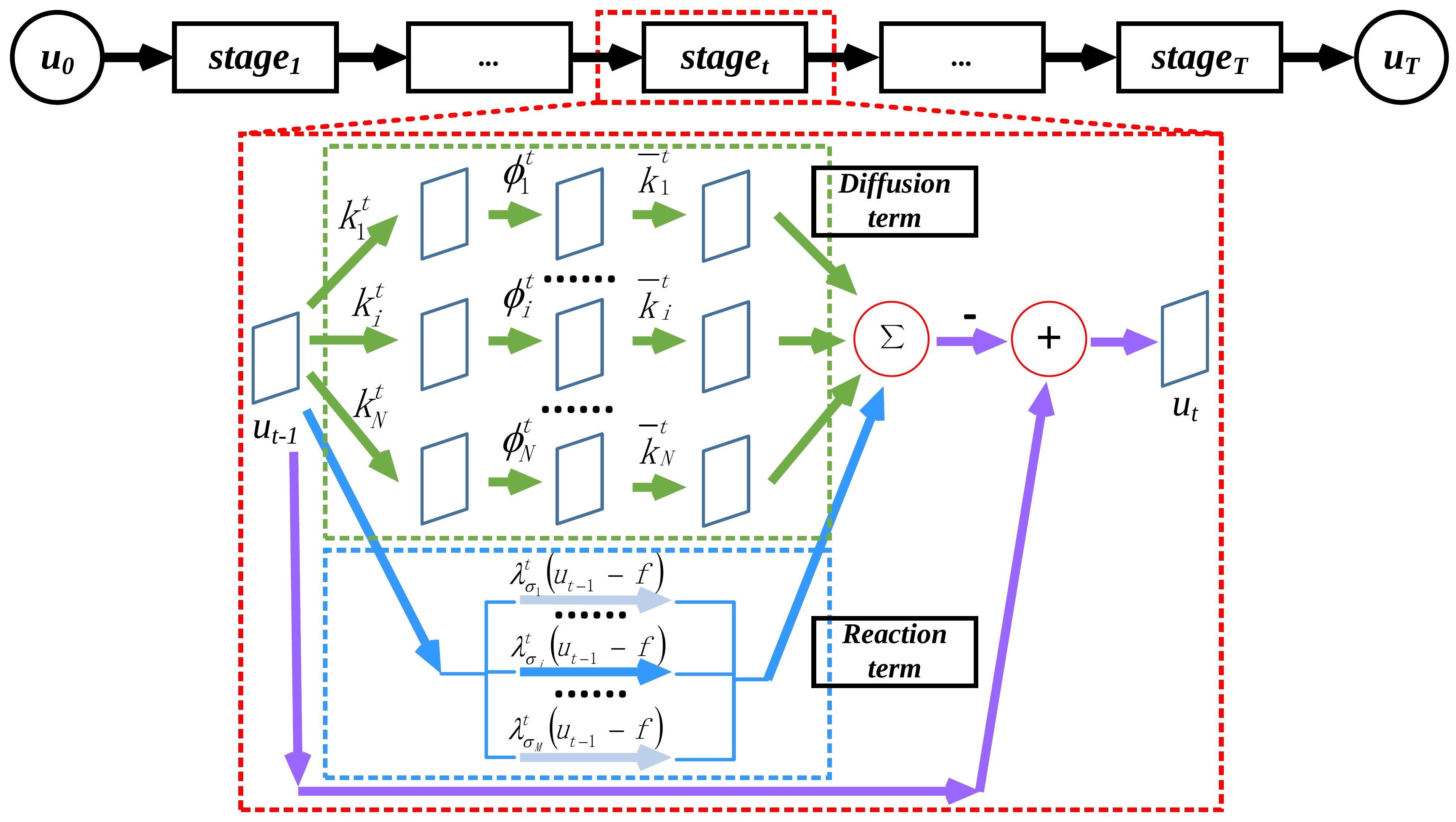}
\caption{The architecture of the proposed genericDP.}\label{fig:networks}
\end{figure}

To handle the inefficiency of discriminative training, we enforce these TNRD models handling different noise level to share one diffusion term, each of which only keeps its own reaction term
\footnote{
By omitting the reaction term, the diffusion process is able to handle multiple Gaussian denoising as well.
Details are discussed in Section \ref{sec:experiment}.
}
, formulated as follow,
\begin{equation}\label{eq:gdp}
\text{E}(u,f_j) = \overset{M}{\underset{i=1}\sum} {\mathds{1}_i \cdot \frac{\lambda_{\sigma_i}}{2} \norm{u-f_i}} + \overset{N_k}{\underset{i=1}\sum} \rho_i(k_i \ast u),
\end{equation}
where $\mathds{1}_i$ is an indicator function that is set to 1 when $i$ = $j$, otherwise is set to 0. $M$ is the number of noise levels. Truncating the gradient descent of minimizing Eq. (\ref{eq:gdp}) with $T$ steps, we arrive the proposed genericDP model, described as follow,
\begin{equation}\label{eq:GDPinferTstep}
\begin{cases}
u_0 = f_j,\quad t = 1, \cdots, T\\
u_t = u_{t-1} -
\left(
\overset{N_k}{\underset{i=1}\sum} {\bar{k}_{i}^{t}} \ast \phi_{i}^{t}(k_{i}^{t} \ast u_{t-1}) +
\lambda_{\sigma_j}^t(u_{t-1} - f_j)
\right) \,.
\end{cases}
\end{equation}
The resulting model is shown in Fig. \ref{fig:networks}.
Given an input $u_0 = f_j$ that degraded by Gaussian noise with $\sigma_j$,
only the data term corresponding to $\sigma_j$ is used.
In this work, we parameterize the local filters and nonlinear functions following \cite{chen2015learning}.

\subsection{Training of GenericDP}
The parameters of the genericDP in (\ref{eq:GDPinferTstep}) are $\Theta = \{\Theta^t\}_{t=1}^{t=T}$, where $\Theta^t = \{k_i^t, \phi_i^t, \lambda_{\sigma_1}^t, \cdots, \lambda_{\sigma_M}^t \}$.
Given these training image pairs of degraded input $f_{j_s}^{s}$ and ground-truth $u_{gt}^s$,
the training procedure is formulated as
\begin{equation}\label{eq:GDPtrain}
\hspace{-0.2cm}\begin{cases}
\Theta^* = \text{argmin}_{\Theta}\cL(\Theta) = \sum\limits_{s = 1}^{S}\ell\left( u_T^s , u_{gt}^s \right) \\
\text{s.t.}
\begin{cases}
u_0^s = f_{j_s}^{s}, \quad t = 1 \cdots T\\
u_t^s = u_{t-1}^s -
\left(
\overset{N_k}{\underset{i=1}\sum} {\bar{k}_{i}^{t}} \ast \phi_{i}^{t}(k_{i}^{t} \ast u_{t-1}^s) +
{\lambda_{\sigma_j}^t}(u_{t-1}^s - f_{j_s}^{s})
\right) ,\\
\end{cases}
\end{cases}
\end{equation}
where $u_T^s$ is the output of the genericDP in Eq. (\ref{eq:GDPinferTstep}).
The inputs are generated using one $\sigma_j$ among $\sigma_1, \cdots, \sigma_M$.
In this paper, the loss function is $\ell(u_T,u_{gt}) = \frac{1}{2} \norm{ u_T - u_{gt} }$.

If not specifically mentioned, we just omit the sample index $s$ to keep the derivation clear.
It is easy to extend the following derivation for one training sample to all training samples.
Using back-propagation technique \cite{lecun1998gradient}, the gradients of $\ell(u_T, u_{gt})$ w.r.t. the parameters $\Theta$ is
\begin{equation}\label{eq:iterstep}
\frac {\partial \ell(u_T, u_{gt})}{\partial \Theta_t} =
\frac {\partial u_t}{\partial \Theta_t} \cdot \frac {\partial u_{t+1}}{\partial u_{t}} \cdots
\frac {\partial \ell(u_T, u_{gt})}{\partial u_T} \,.
\end{equation}

\textbf{Computing the gradient of $\ell(u_T,u_{gt})$ w.r.t $\lambda_{\sigma_j}^t$.}
Given Eq. (\ref{eq:GDPinferTstep}), the derivative of $u_t$ w.r.t. $\lambda_{\sigma_j}^t$ is computed as
\begin{equation}\label{eq:grad_u_lambda}
\frac{\partial u_t}{\partial \lambda_{\sigma_{j}}^t} = -(u_{t-1}-f_j)^{\top}.
\end{equation}

Coining $\frac{\partial \ell(u_T, u_{gt})}{\partial u_t} = e$, the derivative of $\ell(u_T,u_{gt})$ w.r.t. $\lambda_{\sigma_{j}}$ is given as
\begin{equation}\label{eq:grad_l_lambda}
\frac{\partial \ell(u_T,u_{gt})}{\partial \lambda_{\sigma_{j}}^t} = -(u_{t-1}-f_j)^{\top}e\,.
\end{equation}
We only use the samples generated by $\sigma_j$ to compute the gradients $\ell(u_T,u_{gt})$ w.r.t. $\lambda_{\sigma_{j}}$, and update the parameter $\lambda_{\sigma_{j}}^t$ with these gradients.

The derivative of $\ell(u_T,u_{gt})$ w.r.t. $k_i^t$ and ${\phi}_i^t$ are similar to that in \cite{chen2015learning}.
Different from the update of parameter $\lambda_{\sigma_{j}}^t$, all training samples are used to update the parameter $k_i^t$ and ${\phi}_i^t$.

\section{Experimental Results}\label{sec:experiment}
In this section, we first investigate the influence of the parameters, and then compare the trained genericDP models with the state-of-the-art denoising methods. Finally, we transfer the trained diffusion term to the non-blind image deconvolution which is unseen in the training phase, and compare it with the state-of-the-art non-blind deconvolution methods.

\subsection{Training Setups}
The training dataset is constructed over 1000 natural images collected from the Internet.
We randomly cropped 2 regions of size $90 \times 90$ from each image,
resulting in a total of 2000 training images of size $90 \times 90$.

Given pairs of noisy input and ground-truth images, we minimize Eq. (\ref{eq:GDPtrain}) to learn the parameters of the genericDP models with commonly used gradient-based L-BFGS \cite{liu1989limited}.

As TNRD model serves as a strong baseline, we initialize the parameters $\Theta$ using the TNRD model.
We tested the trained models on the 68 images test dataset \cite{roth2009fields}.
We evaluated the denoising performance using PSNR \cite{chen2015learning} and SSIM \cite{Wang04imagequality}.

\subsection{Influence of the Parameters}

\textbf{Number of the training samples.} In this experiment, the inference stage $T$ was set to 8, the range of the noise level $M$ was set to 25.
In \cite{chen2015learning}, 400 images of size $180 \times 180$ are used.
In terms of the number of total pixels, 1600 images of size $90 \times 90$ may be sufficient.
Therefore, we use 2000 images of size $90 \times 90$ as described above.
We also used 4000 images to train the models by doubling each training image in those 2000 images.

We first trained the $\text{TNRD}_{7 \times 7}^8$ model using the new training samples.
The trained TNRD model provided almost the same denoising performance with the models trained using 400 images of size $180 \times 180$ in \cite{chen2015learning}.
Therefore, using the new training images to train the genericDP models does not introduce extra image content, which may contribute to the model performance improvement.

As shown in Table \ref{table:NumSample}, given more training images, the overall performance was improved and was very competing with the original TNRD trained for each specific noise level.
Therefore, $S=4000$ is preferred.

\textbf{Range of the noise levels.} In this experiment, the inference stage $T$ was set to 8, the number of the training samples $S$ was set to 4000.
We investigated the influence of the range of noise level $M$, by shortening the range to $M=15$ and enlarging the range to $M=50$.
In $M=15$, compared with that in $M=25$, the performance was improved only in $\sigma=5$, but very limited .
In $M=50$, compared with that in $M=25$, the performance in $\sigma=5$ was reduced by 0.17dB;
the performance in other $\sigma$ were similar to that in $M=25$.
As the time of training these models with different $M$ is almost the same, $M=50$ is preferred.

Note that, we only trained one genericDP model instead of 50 TNRD models to handle multiple Gaussian denoising with $\sigma$ ranging of 1 to 50.
Therefore, the training time of the genericDP model is almost 50 times faster than that of training 50 TNRD models.

\begin{table}[tp]
\centering
\scriptsize
\caption{Influence of the number of training samples, the range of noise level and with/without reaction term.
}\label{table:NumSample}
\begin{tabular}{|l|c|c|c|c|c|}
\hline
  & \multicolumn{4}{c|}{$\sigma$} \\
\cline{2-5}
{ } & {5} & {15} & {25} & {50} \\
\hline
$\text{TNRD}_{7 \times 7}^8$ & 37.77 & 31.42 & 28.94 & 26.01 \\
\hline
2k-M=25 & 37.57 & 31.38 & 28.85 & - \\
4k-M=25 & 37.55 & 31.41 & 28.91 & - \\
\hline
4k-M=15 & 37.65 & 31.40 & - & - \\
\hline
4k-M=50 & 37.38 & 31.35 & 28.90 & 25.99 \\
\hline
4k-M=50-w/o & 34.19 & 30.78 & 28.55 & 25.68 \\
\hline
\end{tabular}
\end{table}

\textbf{Inference stage.} In \cite{chen2015learning}, the TNRD model is considered as a multi-layer network or convolutional network.
Meanwhile deeper models, \eg, VGG \cite{simonyan2014very} and ResNet \cite{he2016deep}, have achieved success in image classification on ImageNet.
Therefore, it is worth trying more inference stages in the genericDP model as well.

The number of training images $S$ was set to 4000, the range of noise level $M$ was set to 50.
We trained the genericDP model by setting inference stage $T$ to 8, 10 and 16
\footnote{There is no available TNRD models for inference stage $T=10$ and $T=16$, we trained the genericDP models in greedy and joint training scheme from a plain initialization.
}.
As inference stage $T$ increasing, the genericDP model did not provide significantly better denoising performance. Considering the training time and model performance, $T=8$ is prefered.

\subsection{With/without Reaction Term}
Intuitively, it makes scene to train the genericDP model without reaction term. The trained genericDP model in such setting, coined as genericDP-wo, offered inferior denoising results compared with the genericDP model trained with reaction term, as shown in the last two rows of Table \ref{table:NumSample}.
The genericDP-wo model merely contains a pure diffusion process. Therefore, it tends to oversmooth the texture regions and/or produces artifacts in the homogeneous regions.
Therefore, we argue that when training the genericDP model, the reaction term is crucial and is preferred in the following experiments.

\subsection{Image Denoising}
\begin{table}[t]
\centering
\caption{Denoising comparison on test images \cite{roth2009fields}.}\label{table:denoise}
\subfigure{\includegraphics[width=1.0\linewidth]{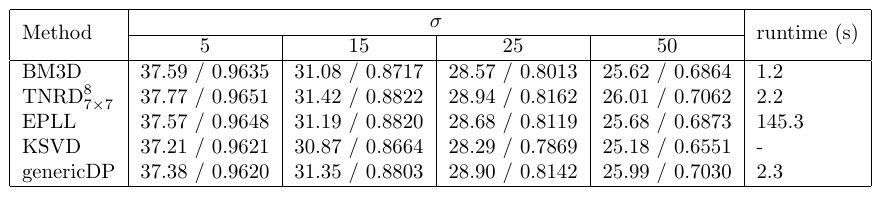}}
\end{table}

\begin{figure}
\centering
\subfigure[original]
{\includegraphics[width=0.24\linewidth]{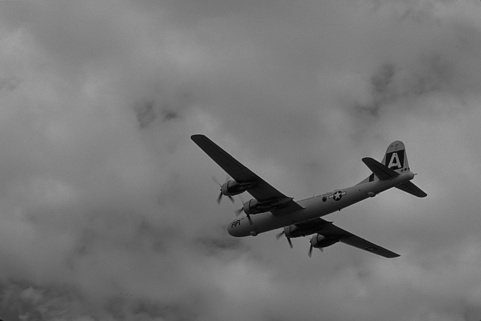}}\hfill
\subfigure[noisy (14.15, 0.0507)]
{\includegraphics[width=0.24\linewidth]{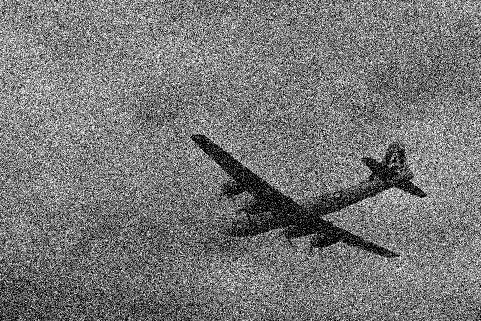}}\hfill
\subfigure[proposed (34.40, 0.9457)]
{\includegraphics[width=0.24\linewidth]{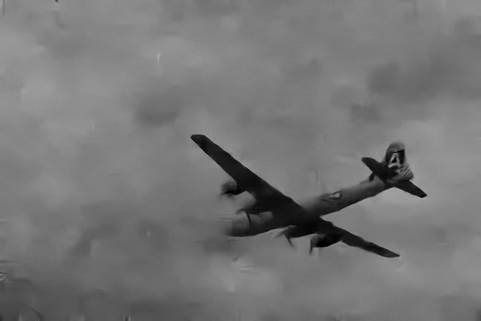}}\hfill
\subfigure[TNRD \cite{chen2015learning} (34.34, 0.9392)]
{\includegraphics[width=0.24\linewidth]{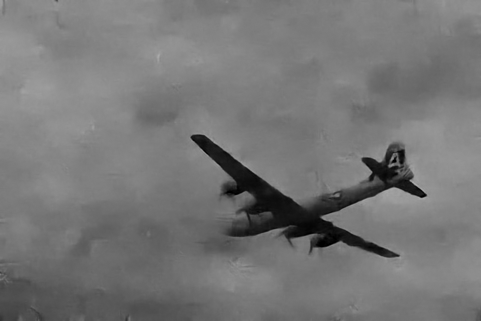}}\\
\subfigure[BM3D \cite{dabov2007image} (33.45, 0.9266)]
{\includegraphics[width=0.24\linewidth]{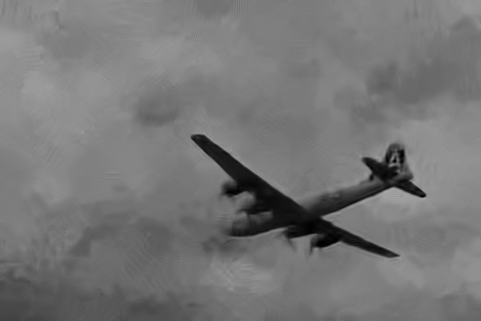}}\hfill
\subfigure[KSVD \cite{aharon2006k} (31.88, 8940)]
{\includegraphics[width=0.24\linewidth]{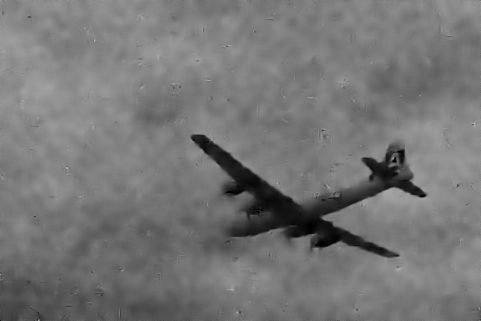}}\hfill
\subfigure[EPLL \cite{zoran2011epll} (32.51, 0.8868)]
{\includegraphics[width=0.24\linewidth]{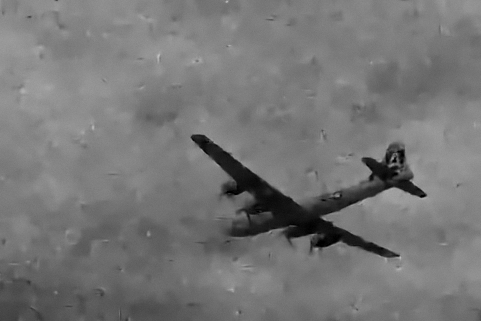}}\\
\caption{
Image denoising comparison.
From left to right, images are generated by original, noisy, ours genericDP model, TNRD, BM3D, KSVD and EPLL.
The original image is added noise level $\sigma=50$.
The numbers in the blankets are PSNR and SSIM values respectively.
}\label{fig:denoise}
\end{figure}

We trained the genericDP model by setting, the number of training images $S=4000$, the range of noise level $M=50$, the inference stage $T=8$, and with reaction term.
We compared the trained genericDP model with BM3D \cite{dabov2007image}, EPLL \cite{zoran2011epll}, KSVD \cite{aharon2006k} and TNRD \cite{chen2015learning} in Gaussian denoising.
The codes of the competing methods were downloaded from the authors' homepage.
The comparison noise level were $\sigma=$5, 15, 25 and 50.
Dealing with one noise level, the genericDP used the corresponding reaction term for that noise level.
The comparison results are listed in Table \ref{table:denoise}.
Visual comparison is shown in Fig. \ref{fig:denoise}.
Despite the efficient transferring, the generative approaches, \eg, EPLL and KSVD, provide inferior denoising performance, and the inference procedure is quite long
\footnote{
The runtime of KSVD varies for different noise level, 761, 117, 51, 18 seconds for $\sigma$ = 5, 15, 25 and 50 respectively.
However, the runtime of KSVD is still larger than that of our genericDP model.
}.
In \cite{rosenbaum2015return}, gating networks are exploited to accelerate the inference process for EPLL-GMM.
While inference time goes down significantly, the gated EPLL provides inferior results compared with EPLL-GMM.

Our genericDP model provides competing results with $\text{TNRD}_{7 \times 7}^8$ trained for a specific noise level,
but runs slightly slower than $\text{TNRD}_{7 \times 7}^8$.

\subsection{Non-blind Deconvolution}
It is more flexible by decouple the data term and regularization term in Eq. (\ref{eq:regularizationModel}) to transfer image prior or prior-like term, \eg, state-of-the-art denoising methods, to other image restoration problems.
\begin{equation}\label{eq:regularizationModel}
E(u,f) = \mathcal{R}(u) + \mathcal{D}(u,f),
\end{equation}
where $\mathcal{R}(u)$ is the regularization term, and
$\mathcal{D}(u,f)$ is the data term.
In \cite{venkatakrishnan2013plug, brifman2016turning, romano2016little, chan2017plug}, Alternating Direction Method of multiplier (ADMM) is exploited to split the regularization and data term,
while in \cite{geman1992constrained, geman1995nonlinear, schmidt2014shrinkage, zhang2017learning} half quadratic splitting (HQS) is used.
In this paper, we focus on HQS approaches.

By introducing a couple of auxiliary variables $z$, one can decouple the data term and regularization term in Eq. (\ref{eq:regularizationModel}), which is reformulated as
\begin{equation}\label{eq:hqs}
\text{E}(u,z,f) = \cR(z) + \cD(Au,f) + \frac{\beta}{2} \norm{u-z},
\end{equation}
where $\frac{\beta}{2} \norm{u-z}$ is the added quadratic term.
$\beta$ is the penalty parameter, when $\beta \rightarrow \infty$, the solution of Eq. (\ref{eq:hqs}) goes very close to that of Eq. (\ref{eq:regularizationModel}).
In HQS setting, Eq. (\ref{eq:hqs}) is divided into two sub-problems as
\begin{equation}\label{eq:hqs1}
\begin{cases}
z_{t+1} = \mathop{\arg\min\limits_{z}} \cR(z) + \frac{\beta}{2} \norm{u_t-z} \\
u_{t+1} = \mathop{\arg\min\limits_{u}} \cD(Au,f) + \frac{\beta}{2} \norm{u-z_{t+1}}.
\end{cases}
\end{equation}
By alternatively minimizing these two sub-problems and increasing $\beta$ iteratively, we can get the estimation of latent image $\hat{u}$.
Sub-problem $z_{t+1}$ can be regarded as a denoising process using image prior $\cR(z)$, \eg, FoE, or using state-of-the-art denoising methods, \eg, BM3D or TNRD.
Sub-problem $u_{t+1}$ aims to find a solution which satisfies the data term and is close to the $z_{t+1}$.
The commonly used data term is in $\ell 2$ norm, \ie, $\cD(Au,f) = \frac{\lambda}{2} \norm{Au - f}$.
With the updated $z_{t+1}$, $u_{t+1}$ sub-problem has a closed-form solution,
\begin{equation}\label{eq:ucf}
u_{t+1} = (\lambda_1 A^{\top}A + \text{I})^{-1}(\lambda_1 A^{\top}f + z_{t+1}),
\end{equation}
where $\lambda_1 = \frac{\lambda}{\beta}$, $\text{I} \in \cR^{p \times p}$ is an identity matrix.

\begin{table}[tp]
\centering
\caption{Non-blind image deconvolution comparison on test images \cite{levin2009understanding}.}\label{table:deconv32}
\subfigure{\includegraphics[width=1.0\linewidth]{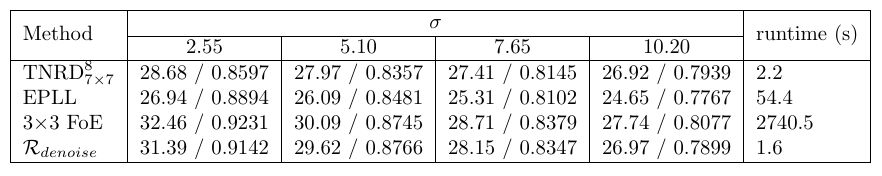}}
\end{table}

The regularization term trained in previous subsection is donated as $\cR_{denoise}$.
Therefore, the optimization of non-blind deconvolution is reformulated as
\begin{equation}\label{eq:hqs2}
\begin{cases}
z_{t+1} = \mathop{\arg\min\limits_{z}} \cR_{denoise}(z) + \frac{\beta}{2} \norm{u_t-z} \\
u_{t+1} = \mathop{\arg\min\limits_{u}} \cD(Hu,f) + \frac{\beta}{2} \norm{u-z_{t+1}},
\end{cases}
\end{equation}
where $\beta$ is increasing exponentially.
Matrix $H$ is the matrix form of the blur kernel $h$.
In this process, $\lambda$ in $D(Hu, f)$ is set to $\frac{100}{1}$, $\frac{100}{4}$, $\frac{100}{9}$ and $\frac{100}{16}$ for $\sigma=$ 2.55, 5.10, 7.65 and 10.20 respectively.
$\beta$ increases exponentially as $\beta = \gamma^{i}$,
where $\gamma$ is set to 1.8, 1.7, 1.6 and 1.5 for $\sigma=$ 2.55, 5.10, 7.65 and 10.20 respectively.
The power $i = t-1$ for each inference stage $t$ respectively.
The number of iteration here is simply the number of inference stage $T$, while the number of iteration is 30 for \cite{zhang2017learning} and more than 100 for \cite{brifman2016turning}.

We compare the $\cR_{denoise}$ with TNRD \cite{chen2015learning}, EPLL \cite{zoran2011epll} and FoE \cite{schmidt2011bayesian} in non-blind deconvolution.
The codes of the competing methods were downloaded from the authors' homepage
\footnote{
There is no available TNRD codes for non-blind deconvolution, we implement it and train it using greedy training.
}.
The test images are from \cite{levin2009understanding}, which are widely used in image deconvolution.
In this test dataset, there are eight blur kernels and four images.
The blurry images are generated in the following way,
firstly applying a blur kernel and then adding zero-mean Gaussian noise with noise level $\sigma$.
The noise levels are 2.55, 5.10, 7.65 and 10.20.

\begin{figure}
\centering
\subfigure[original]
{\includegraphics[width=0.16\linewidth]{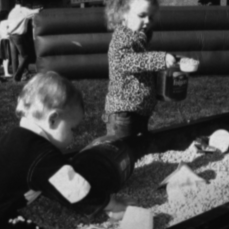}}\hfill
\subfigure[blurred (17.30, 0.3917)]
{\includegraphics[width=0.16\linewidth]{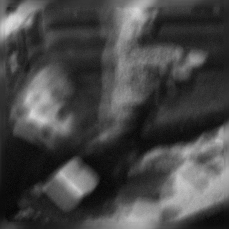}}\hfill
\subfigure[proposed (24.43, 0.8131)]
{\includegraphics[width=0.16\linewidth]{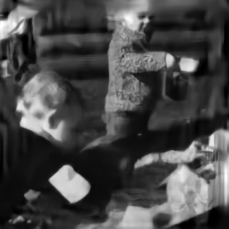}}\hfill
\subfigure[TNRD \cite{chen2015learning} (24.36, 0.7860)]
{\includegraphics[width=0.16\linewidth]{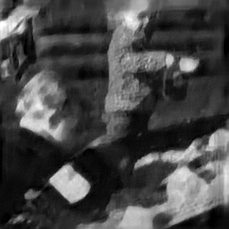}}\hfill
\subfigure[FoE \cite{schmidt2011bayesian} (30.16, 0.8922)]
{\includegraphics[width=0.16\linewidth]{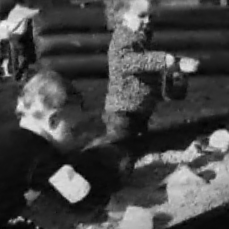}}\hfill
\subfigure[EPLL \cite{zoran2011epll} (21.59, 0.7986)]
{\includegraphics[width=0.16\linewidth]{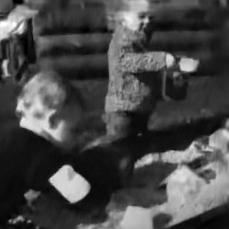}}\\
\subfigure[original]
{\includegraphics[width=0.16\linewidth]{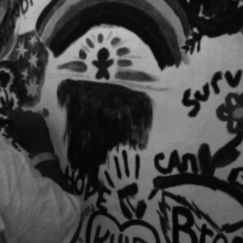}}\hfill
\subfigure[blurred (24.79, 0.7514)]
{\includegraphics[width=0.16\linewidth]{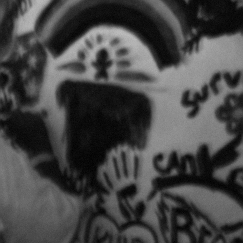}}\hfill
\subfigure[proposed (36.73, 0.9661)]
{\includegraphics[width=0.16\linewidth]{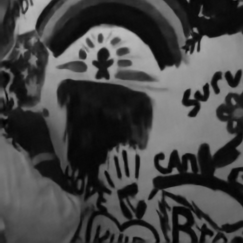}}\hfill
\subfigure[TNRD \cite{chen2015learning} (32.76, 0.9299)]
{\includegraphics[width=0.16\linewidth]{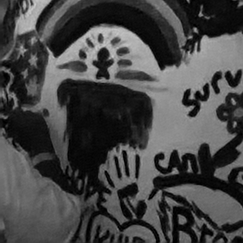}}\hfill
\subfigure[FoE \cite{schmidt2011bayesian} (35.32, 0.9580)]
{\includegraphics[width=0.16\linewidth]{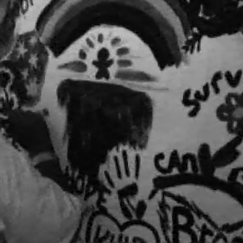}}\hfill
\subfigure[EPLL \cite{zoran2011epll} (29.67, 0.9528)]
{\includegraphics[width=0.16\linewidth]{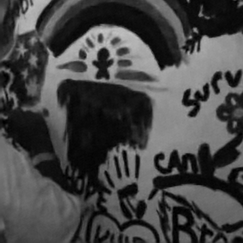}}\\
\subfigure[original]
{\includegraphics[width=0.16\linewidth]{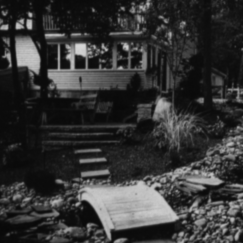}}\hfill
\subfigure[blurred (22.39, 0.4478)]
{\includegraphics[width=0.16\linewidth]{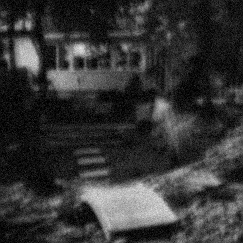}}\hfill
\subfigure[proposed (29.41, 0.8200)]
{\includegraphics[width=0.16\linewidth]{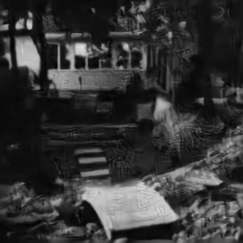}}\hfill
\subfigure[TNRD \cite{chen2015learning} (28.50, 0.8148)]
{\includegraphics[width=0.16\linewidth]{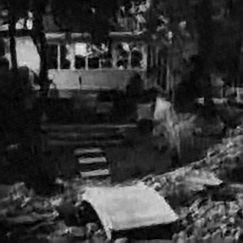}}\hfill
\subfigure[FoE \cite{schmidt2011bayesian} (28.20, 0.8205)]
{\includegraphics[width=0.16\linewidth]{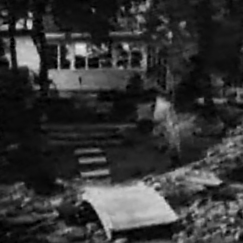}}\hfill
\subfigure[EPLL \cite{zoran2011epll} (26.60, 0.7987)]
{\includegraphics[width=0.16\linewidth]{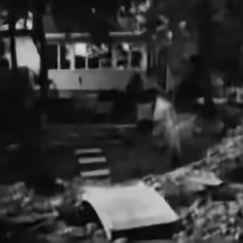}}\\
\caption{
Non-blind image deconvolution comparison.
From left to right, images are generated by original, blurred, ours genericDP model, TNRD, FoE and EPLL.
In the first two rows, the original image is blurred by kernel 4 and 5 respectively, then is added noise level $\sigma=2.55$.
In the last row, the original image is blurred by kernel 5, then is added noise level $\sigma=10.20$.
The numbers in the blankets are PSNR and SSIM values respectively.
}\label{fig:deconv}
\end{figure}

As illustrated in Table \ref{table:deconv32}, the $\cR_{denoise}$ provides competing results with FoE, and better results than EPLL and TNRD.
The $\cR_{denoise}$ runs slightly faster than TNRD
\footnote{
Using HQS, we can accelerate the sub-problem $u_{t+1}$ described in Eq. \ref{eq:hqs2} using Fast Fourier Transform (FFT). Note that one needs to careful handle the image boundary conditions.}, and faster than EPLL and FoE.
While FoE method provides good deconvolution results, the inference time is very long.
Therefore, it is not scalable to recover large image for FoE.
Considering the inference efficiency and deconvolution performance, the $\cR_{denoise}$ is very promising.
Visual comparison is shown in Fig. \ref{fig:deconv}.

\section{Conclusion}
\label{sec:conclusion}

Instead of training multiple TNRD models, we enforce these diffusion processes sharing one diffusion term and keeping its own reaction term.
As a result, we only need to train one model to handle multiple Gaussian denoising of $\sigma$ in a range.
We derive the gradients of loss function w.r.t. the parameters, and train the model in a supervised and end-to-end manner.
The trained genericDP model can offer very competing denoising performance compared with the original TNRD model trained for each specific noise level.
Meanwhile, the training efficiency is very impressive compared with TNRD and even generative approaches.
We transfer the trained diffusion term to non-blind deconvolution using HQS method.
Experiment results show that the trained diffusion term can be used as a generic image prior and work well in image non-blind deconvolution which is unseen during training.

In this work, we train the genericDP model using images only degraded by Gaussian noise in a range.
We will use more types of degradation operators $A$, \eg, image super-resolution, image deblocking, to train the genericDP model, hoping that a more generic image prior can be learned.
We will also transfer the learned diffusion term to other unseen image restoration problems to validate the generality of the trained diffusion term.

\section{Acknowledgements}
This work was supported by
the National Natural Science Foundation of China
under the Grant No.U1435219, No.61732018 and No.61602032.
\bibliography{egbib}
\end{document}